\documentclass[letterpaper]{article} 
\usepackage{aaai25}  
\usepackage{times}  
\usepackage{helvet}  
\usepackage{courier}  
\usepackage[hyphens]{url}  
\usepackage{graphicx} 
\usepackage{natbib}  
\usepackage{caption} 
\usepackage{algorithm}

\usepackage{newfloat}
\usepackage{listings}
\DeclareCaptionStyle{ruled}{labelfont=normalfont,labelsep=colon,strut=off} 
\floatstyle{ruled}
\newfloat{listing}{tb}{lst}{}
\floatname{listing}{Listing}
\newcommand{\method}{Compatibility-aware Knowledge Integration}
\newcommand{\methodbrief}{CKI}
\newcommand{\moduleAfull}{Parameter Compatibility Assessment}
\newcommand{\moduleA}{Parameter Uncertainty Assessment}
\newcommand{\moduleAAA}{Model Information Content Assessment}
\newcommand{\moduleB}{Parameter Splicing}

\usepackage{todonotes}
\renewcommand{\todo}[1]{\iffalse #1 \fi{\color{blue} \textbf{[TODO]}}}

\newcommand{\leftrarrows}{\mathrel{\raise.9ex\hbox{\oalign{%
  $\scriptstyle\leftarrow$\cr
  \vrule width0pt height.5ex$\hfil\scriptstyle\relbar$\cr}}}}
\newcommand{\lrightarrows}{\mathrel{\raise.9ex\hbox{\oalign{%
  $\scriptstyle\relbar$\hfil\cr
  $\scriptstyle\vrule width0pt height.5ex\smash\rightarrow$\cr}}}}
\newcommand{\Rrelbar}{\mathrel{\raise.9ex\hbox{\oalign{%
  $\scriptstyle\relbar$\cr
  \vrule width0pt height.5ex$\scriptstyle\relbar$}}}}

\usepackage{amssymb}
\usepackage{amsmath}
\usepackage{enumitem}
\usepackage{newfloat}
\usepackage{multirow}
\usepackage{makecell}
\usepackage{xcolor}

\usepackage{colortbl}
\usepackage{dsfont}
\usepackage{enumitem}
\usepackage{algpseudocode}
\usepackage{tcolorbox} 
\usepackage{amsthm}
\theoremstyle{plain}

\theoremstyle{definition}

\theoremstyle{remark}

\usepackage{cleveref}       

\definecolor{myyellow}{rgb}{1,1, 0.6}
\definecolor{myorange}{rgb}{1, 0.8, 0.6}
\definecolor{myred}{rgb}{1, 0.6, 0.6}
\definecolor{second}{HTML}{FFDAB9}
\definecolor{best}{HTML}{FFC1C1}
\usepackage{multirow}
\usepackage{colortbl}
\usepackage{bbding}
\usepackage{stackrel}
\usepackage{booktabs}
\DeclareMathOperator*{\argminA}{arg\,min} 

\title{Optimize Incompatible Parameters \\ Through Compatibility-aware Knowledge Integration}

\author{Zheqi Lv$^{1,3}$, Keming Ye$^{1}$, Zishu Wei$^{1}$, Qi Tian$^{2}$, Shengyu Zhang$^{1}$\thanks{Corresponding authors.}, Wenqiao Zhang$^{1}$,\\Wenjie Wang$^{3*}$, Kun Kuang$^{1*}$, Tat-Seng Chua$^{3}$, Fei Wu$^{1}$
}
\affiliations{
\textsuperscript{$1$}Zhejiang University, Hangzhou, China \\
\textsuperscript{$2$}Tencent TEG., Shenzhen, China \\
\textsuperscript{$3$} National University of Singapore, Singapore \\
\{zheqilv, kemingye, weizishu, sy\_zhang, wenqiaozhang, kunkuang, wufei\}@zju.edu.cn, \\noaltian@tencent.com, wenjiewang96@gmail.com, dcscts@nus.edu.sg
}

\usepackage{bibentry}

\begin{document}

\maketitle

\begin{abstract}
\label{sec:abstract}
Deep neural networks have become foundational to advancements in multiple domains, including recommendation systems, natural language processing, and so on. Despite their successes, these models often contain incompatible parameters that can be underutilized or detrimental to model performance, particularly when faced with specific, varying data distributions. Existing research excels in removing such parameters or merging the outputs of multiple different pretrained models. However, the former focuses on efficiency rather than performance, while the latter requires several times more computing and storage resources to support inference.
In this paper, we set the goal to explicitly improve these incompatible parameters by leveraging the complementary strengths of different models, thereby directly enhancing the models without any additional parameters. Specifically, we propose \textbf{\method{}} (\textbf{\methodbrief{}}), which consists of \moduleAfull{} and \moduleB{}, which are used to evaluate the knowledge content of multiple models and integrate the knowledge into one model, respectively. The integrated model can be used directly for inference or for further fine-tuning. 
Extensive experiments on various recommendation and language datasets show that \methodbrief{} can effectively optimize incompatible parameters under multiple tasks and settings to break through the training limit of the original model without increasing the inference cost.
\end{abstract}

\section{Introduction}
\label{sec:introduction}

Deep neural networks have excelled across various domains including natural language processing~\cite{ref:liu2019roberta,ref:devlin2018bert}, recommender systems~\cite{ref:gru4rec,ref:sasrec}, etc. The work on the lottery ticket hypothesis~\cite{ref:frankle2020linear,ref:frankle2023lottery} shows that there exist subnetworks that can be trained in isolation to achieve full accuracy. However, the parameters except these subnetworks, when applied to specific data distributions, often contain parameters that contribute minimally or even negatively to task performance~\cite{ref:init_survey,ref:init_pruning_survey}. These parameters can be referred to as ``incompatible parameters''.

Existing methods to tackle the negative impact of these parameters primarily include: 1) Removing these incompatible parameters (such as pruning and so on)~\cite{ref:init_pruning_survey,ref:pruning_ma2023llm,ref:pruning_guo2020multi,ref:pruning_zhou2023three,ref:pruning_han2015learning,ref:pruning_wang2020pruning,ref:pruning_dong2017learning,ref:pruning_sanh2020movement} to  enhance computational efficiency is well-established. However, in dynamic environments, characterized by shifting data distributions (\textit{e.g.}, context and scenarios on language tasks, user preferences on recommendation tasks.), pruning may lose its efficiency. 2) An alternative strategy is training multiple models under different conditions (e.g., with different random seeds and so on) and merging their outputs during inference (such as outputs ensemble~\cite{ref:ensemble_dong2020survey,ref:ensemble_zhou2021domain_adaptation}), which increase accuracy by aggregating the outputs of several models trained under different conditions. This approach can mitigate the negative effects of incompatible parameters by leveraging complementary strengths from diverse models. However, this benefit comes at the cost of significantly higher computational demands during inference, and it does not address the core issue of improving the incompatible parameters themselves.

\begin{figure*}
    \centering
    \includegraphics[width=\textwidth]{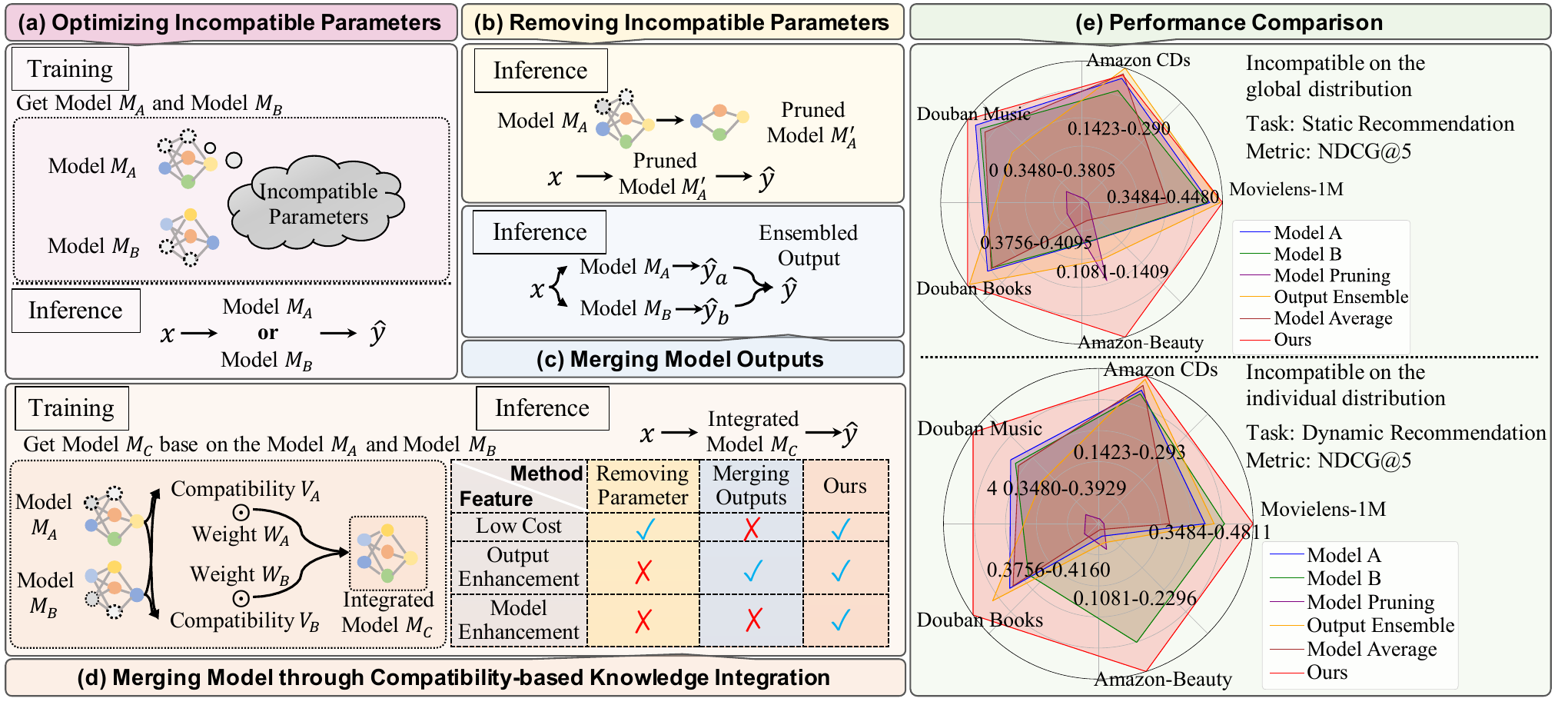}
    \caption{
    (a) shows the Incompatible Parameter issue.
    (b) describes Model Pruning, which removes incompatible parameters from $M_A$.
    (c) presents output ensemble, which combines the inference results of $M_A$ and $M_B$ for a final result.
    (d) introduces \methodbrief{}, which evaluates each parameter's compatibility in global and local views, then integrates the knowledge of $M_A$ and $M_B$ to get model $M_C$.
    (e) shows that \methodbrief{} outperforms baselines in different scenarios.
    }
    \label{fig:introduction}
\end{figure*}

Inspired by the complementary model capabilities in output ensemble, we propose a novel perspective in optimizing incompatible parameters, that is identifying these parameters and integrating complementary counterparts from models trained under diverse conditions for enhancement.
Our goal is to establish a unified framework for knowledge integration, applicable to multiple domains including recommendation and language tasks. This integrative approach acknowledges the ubiquitous presence of potentially complementary models in real-world applications. For instance, within large online platforms such as Amazon and Taobao, models dedicated to understanding multi-modal content and user behaviors at different stages, such as initial engagement, search, and post-purchase, can provide mutually reinforcing insights.
A straightforward strategy might involve averaging the parameters across models to synthesize a unified model for inference. However, this approach can fall short of expectations, as illustrated in Figure~\ref{fig:introduction}.

Specifically, we design a novel Unified \method{} (\methodbrief{}) algorithm, to address the aforementioned limitations. The core idea of our approach, is to assess the compatibility at each parameter position of each model, and then fuse the parameter across multiple models based on these uncertainties.
As shown in Figure~\ref{fig:introduction}, \methodbrief{} comprises the following parts: 
(1) \textbf{\moduleAfull{}} includes two aspects: local-level parameter uncertainty to assess the uncertainty of each parameter position, and global-level model information content to assess the significance of the entire model. Subsequently, \moduleAfull{} fuses the local-level model uncertainty and global-level model information content into a comprehensive parameter compatibility. This dual-perspective framework is predicated on the understanding that the fidelity of model parameters is not uniformly distributed across the model's architecture; discrepancies can arise both at the granularity of individual parameters and at the holistic level of the model's overall parameter configuration. This distinction underpins our methodology, which synergistically integrates these two evaluative lenses into a unified parameter compatibility metric.
\emph{Local-level Uncertainty Assessment} examines the variance in parameter values across analogous positions within different model instances. This analysis hinges on the premise that the consistency—or lack thereof—in parameter values at a given position serves as an indicator of the reliability and uncertainty of those parameters. For example, minimal variance among models in a specific parameter suggests a convergence towards a stable, and presumably optimal, value, whereas an outlier may signal a deviation from this collective wisdom.
\emph{Global-level Information Content Assessment} converts model parameters into histogram distribution, and then further obtains information entropy, which is also the basis for measuring the global-level information content of the model or the global-level information content measurement.
We posit that a model's information entropy—a measure of the diversity and richness of information encapsulated within—correlates with the significance of its parameters. A model distinguished by high information entropy is posited to possess a more robust and nuanced parameter set, capable of capturing a broader spectrum of features and nuances within the training data.
\emph{Dual-Perspective Compatibility Blend} adopts a holistic evaluation of parameter compatibility that encapsulates both individual parameter efficacy and the collective synergy of the model's parameters. This comprehensive approach enables a more nuanced understanding of model compatibility, facilitating the identification of models that not only perform well in aggregate but are also constructed from high-compatibility components at every level of their architecture.
(2) \textbf{\moduleB{}} involves splicing the parameters of multiple models based on the parameter compatibility. 
The information entropy of multiple models at each position is extracted nonlinear features by neural networks to be used as splicing weights. To splice the parameters, we design hard splicing and soft splicing methods. \emph{Hard Splicing} adopts a decisive approach, selecting for each parameter position the most optimal parameter from the pool of pre-trained models based on compatibility assessment. This method ensures that the spliced model is made up of the best parts available, albeit with the caveat of potential information loss due to the outright exclusion of lower-compatibility parameters.
To mitigate this, we introduce \emph{Soft Splicing}, which calculates the  weights of the spliced model under the guidance of the weighted sum of parameter compatibility and parameter weights, maximizing the preservation of parameter information and optimizing model parameters.

It should be noted that, the integrated model has the same structure and the same number of parameters as the original pre-trained models, so it does not add even a little bit of additional inference cost. Our \methodbrief{} algorithm has a very wide range of applications. 
It can optimize parameters that are incompatible with the global data distribution as well as those that are incompatible with the individual data distribution. 
The integrated model can be used for inference without any retraining. Furthermore, the integrated model can serve as a better initialization model, requiring only one epoch of retraining to achieve better performance.

In summary, our contributions can be summarized as:
\begin{itemize}[itemsep=0pt,topsep=2pt,leftmargin=10pt]
\item We focus on optimizing incompatible parameters, which is under-explored but important, with the goal of directly enhancing the model without any additional parameters.

\item We instantiated the incompatible parameter optimization method and designed a universal \methodbrief{} framework consists of \moduleAfull{} and \moduleB{}. 

\item We conducted comprehensive and extensive experiments on recommendation and language tasks. The rich experimental results prove the effectiveness and universality of our method.
\end{itemize}

\section{Related Work}
\label{sec:related_work}
\noindent \textbf{Incompatible Parameters.}
Deep learning models often suffer from excessive parameters and redundancies. Network pruning~\cite{ref:pruning_guo2020multi,ref:pruning_wang2020pruning,ref:pruning_sanh2020movement,ref:pruning_han2015learning,ref:pruning_dong2017learning,ref:pruning_ma2023llm,ref:pruning_zhou2023three,ref:pruning_fang2024structural} aims to eliminate non-essential parameters to maintain effectiveness. Evaluation methods include second-order derivatives of the loss function, parameter magnitudes, and other metrics. Output Ensemble~\cite{ref:ensemble_lee2021sunrise_rl,ref:ensemble_zhou2021domain_adaptation,ref:ensemble_kumar2023power,ref:ensemble_mazari2024bert,ref:ensemble_yuan2023dynamic,engine,graphbridge} aggregates results from multiple models trained under different conditions, optimizing final output without altering model parameters.

\noindent \textbf{Model Aggregation.}
Model Aggregation is widely used in federated learning~\cite{ref:federated_fedavg,ref:federated_multi_task,ref:federated_multi_task2,ref:federated_1,ref:federated_2,ref:federated_3,ref:federated_4,ref:federated_5}. Typically, models are trained on devices, with parameters or gradients sent to the cloud, then models are aggregated and send back. Federated learning aims to match the performance of a globally trained model under privacy constraints, which is different from our target, its performance upper bound is a global model's performance. Model Grafting~\cite{ref:model_grafting,ref:model_grafting2} also supports model aggregation but requires pre-trained models on multiple datasets and has different objectives.

\noindent\textbf{Deep Learning Models.}
Deep learning models have achieved excellent performance in various tasks. In vision tasks, models such as ResNet~\cite{ref:he2016resnet}, SqueezeNet~\cite{ref:squeezenet}, MobileNet~\cite{ref:mobilenet,ref:mobilenetv2,ref:mobilenetv3} have significantly improved efficiency. In language tasks, GPT~\cite{ref:gpt1,ref:gpt2,ref:gpt3}, BERT~\cite{ref:devlin2018bert}, and Roberta~\cite{ref:liu2019roberta,ref:devlin2018bert} excel through pre-training and attention mechanisms. In recommendation tasks, models like DeepFM~\cite{ref:deepfm}, LightGCN~\cite{ref:he2020lightgcn}, DIN~\cite{ref:din}, GRU4Rec~\cite{ref:gru4rec}, SASRec~\cite{ref:sasrec}, Bert4Rec~\cite{ref:bert4rec}, and ICLRec~\cite{ref:chen2022iclrec} are often used. Dynamic neural networks are now widely used in computer vision~\cite{alaluf2022hyperstyle,dinh2022hyperinverter}, recommendation systems~\cite{lv2024semantic,yan2022apg}, large language models~\cite{zhang2024hyperllava}, device-cloud collaboration~\cite{ref:duet,lv2024intelligent}, and other fields. They can adjust model parameters based on samples, enabling model personalization.
\section{Methodology}
\label{sec:method}
\subsection{Problem Formulation and Notations}
First, we need to introduce the notation in deep learning. We use $\mathcal{X}=\{x\}$ to represent a piece of data, and $\mathcal{Y}=\{y\}$ to represent the corresponding label. We represent the dataset as $\mathcal{D}$, where $\mathcal{D} = \{X, Y\}$. More specifically, we use $\mathcal{D}_{\rm{Train}}$ to represent the training set and $\mathcal{D}_{\rm{Test}}$ to represent the test set. Roughly speaking, let $\mathcal{L}$ be the loss obtained from training on dataset $\mathcal{D}_{\rm{Train}}$. Then, the model parameters $W$ can be obtained through the optimization function $\argminA \mathcal{L}$. The symbol ``:='' denotes assigning the value on the right side to the left side. Here, we formalize Model Pruning, Output Ensemble, and \methodbrief{} to more clearly demonstrate the differences among these three approaches.

\noindent \textit{Model Pruning.} involves trimming the less important model parameters after training and obtaining the model parameters $W$. There are various ways of pruning, and if we assume $M$ is a mask matrix, these pruning algorithms can be roughly formalized as:
\begin{equation}
    W := W \odot M
\end{equation}

\noindent \textit{Output Ensemble.} assumes that based on the dataset $\mathcal{D}_{\rm{Train}}$, $n$ models are trained (where $n\geq2$). For simplicity, let's initially set $n$ to 2 and assume that the outputs of the two trained models for the same sample $x$ are $\hat{y}_A$ and $\hat{y}_B$, respectively. Then, Output Ensemble can be formalized as:
\begin{equation}
    \hat{y} = \alpha \cdot \hat{y}_A + \beta \cdot \hat{y}_B
\end{equation}

\begin{figure*}
    \centering
    \includegraphics[width=\textwidth]{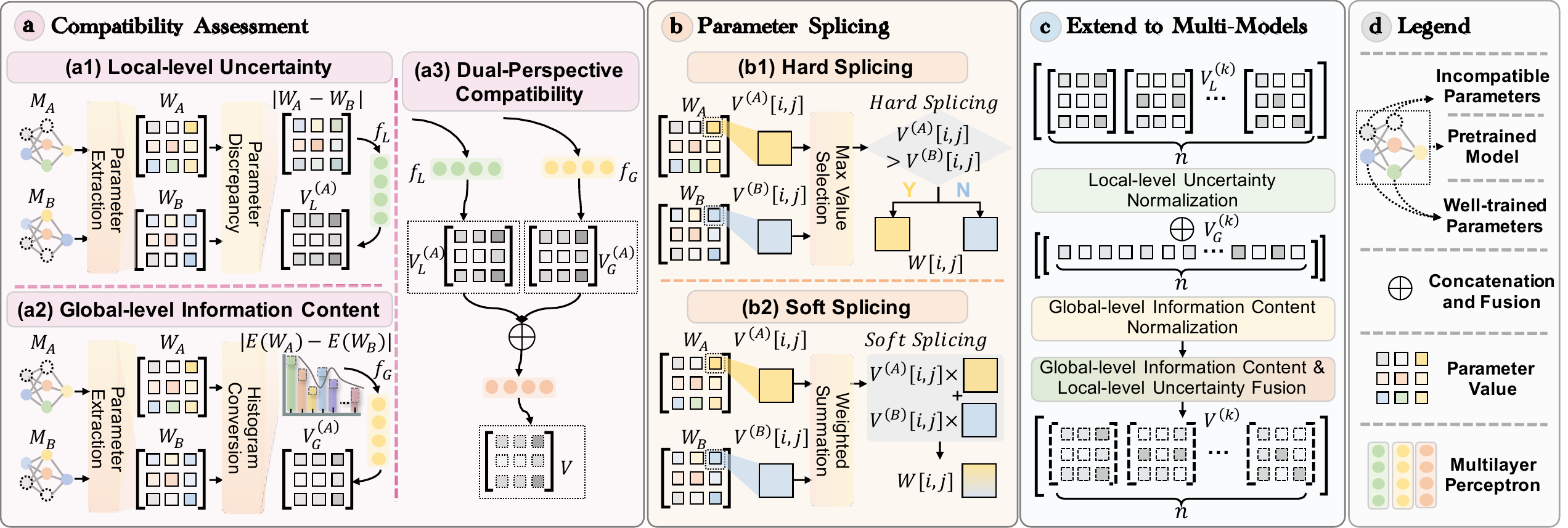}
    \caption{Overview of the proposed \methodbrief{}. Our \methodbrief{} includes two parts: \moduleAfull{} and \moduleB{}. (a) describes the \moduleAfull{}. It consists of 3 parts: (a1) Local-level \moduleA{}, (a2) Global-level \moduleAAA{}, and (a3) Dual-Perspective \moduleAfull{}. (b) describes the \moduleB{}, which includes  (b1) Hard Splicing and (b2) Soft Splicing. (c) describes the extension of \methodbrief{} from 2 models to multiple models.}
    \label{fig:method}
\end{figure*}

\noindent \textit{\methodbrief{}.} CKI also assumes that based on the dataset $\mathcal{D}_{\rm{Train}}$ like Output Ensemble, $n$ models are trained (where $n \geq 2$). For simplicity, let's initially set $n$ to 2 and assume that the parameters of the two trained models are $W_A$ and $W_B$. Similar to Model Pruning, let's assume that the uncertainty matrices for $W_A$ and $W_B$ are $V^{(A)}$ and $V^{(B)}$, respectively. Then, \methodbrief{} can be formalized as,
\begin{equation}
    W= W_A \odot V^{(A)} + W_B \odot V^{(B)}
\end{equation}

\subsection{\methodbrief{} for Dual Models}
In this section, we will introduce our \methodbrief{}. As shown in Figure~\ref{fig:method}, our \methodbrief{} includes two parts: \moduleAfull{} and \moduleB{}. (a) Describes the \moduleAfull{}. It consists of three parts: (a1) describes the Local-level \moduleA{} method, which is used to measure the uncertainty of each parameter in each model. (a2) describes the Global-level \moduleAAA{}, which is used to measure the model information content from a global view. (a3) describes the Dual-Perspective \moduleAfull{}, which assesses model uncertainty from both local-level and global-level perspectives. (b) Describes the \moduleB{}, which includes two parts: (b1) describes the hard splicing, where low-compatibility parameters at each position directly replace high-compatibility parameters from two models. (b2) describes the soft splicing, where each parameter is weighted and fused based on the compatibility of the parameter at that position in the two models. (c) describes the extension of \methodbrief{} from 2 models to multiple models. The details are elaborated in the following subsections.
\subsubsection{\moduleAfull{}}
Here we introduce the methods for assessing parameter compatibility, including Local-level \moduleA{}, and Global-level \moduleAAA{}.

\noindent \textit{Local-level \moduleA{}.}
In the Local-level \moduleA{}, we focus on the uncertainty calculation of each parameter.
The basic paradigm is to train a \moduleA{} network $f$, inputting the network parameters to directly obtain the uncertainty of the parameters. Assuming the dimension of the parameter matrix $W\in\mathbb{R}^{N_{\text{row}} \times N_{\text{col}}}$ is $ N_{\text{row}} \times N_{\text{col}} $, then the dimension of the resulting parameter uncertainty matrix $V\in\mathbb{R}^{N_{\text{row}} \times N_{\text{col}}}$ will also be $ N_{\text{row}} \times N_{\text{col}} $. We use $\phi$ to represent a series of transformations, then the basic paradigm can be formulated as $ V = f(\phi(W))$.
The difference in parameters of multiple models obtained from the same data is important for judging the uncertainty of the parameters at a position. An intuitive example is, if the parameter values at the same position in two models are equal, then the uncertainty of these two parameters may be considered little. If the parameter values at the same position in two models are not equal, then the uncertainty of these two parameters may be considered large. The uncertainty formula is as follows,
\begin{equation}
    V_L^{(A)}[i,j] = f_{L}(|W_A[i,j] - W_B[i,j]|).
\label{eq:local_quality}
\end{equation}
$V_L^{(A)}$ represents the local-level parameter uncertainty matrix of the model $M_A$. 

\noindent \textit{Global-level \moduleAAA{}.}
Although Local-level \moduleA{} can measure the uncertainty of each parameter at its position, it lacks a global perspective. We need to focus on the global perspective to determine which model should be given more credence during model \moduleB{}, and what the weights should be.
Therefore, we further designed Global-level \moduleAAA{}. This concept primarily draws inspiration from the design philosophies of CNNs and transformers. CNNs use a certain receptive field to continuously extract features, enabling smaller feature maps to have a global perspective. 
Similarly, transformers require that at each stage, not only local-level features but also global features need to be extracted.
However, the global-level information content cannot be directly calculated because the values are too discrete. Therefore, we first obtain the parameter distribution histogram based on the parameter values and then derive the information entropy from the parameter distribution histogram to quantify the information content of a model from a global perspective. 
We divide the range between the minimum and maximum values of the model parameters into $u$ equal parts. Then, we can determine the interval for each segment and define the lower bound and upper bound of the $t$-th according to the following formula.
\begin{equation}
\left\{
    \begin{aligned}
    L_t &= W_{\text{min}} + \frac{(W_{\text{max}} - W_{\text{min}})(t - 1)}{u}, \\
    U_t &= W_{\text{min}} + \frac{(W_{\text{max}} - W_{\text{min}})t}{u}.
    \end{aligned}
\right.
\end{equation}
In the above expression, \( W_{\text{max}} \) and \( W_{\text{min}} \) represent the maximum and minimum values of all the parameters in the parameter matrix, respectively. Subsequently, we count the number $c_t$ of parameters falling into each segment to get the probability $p$ of the parameters falling within a distribution.
\begin{equation}
    c_t = \#\{ w \in W \mid w \in [L_t, U_t] \}.
\end{equation}
From this, we can calculate the information entropy  $E(W)$ of the parameters $W$ based on $c_t$.
\begin{equation}
    E(W) = -\sum_{i=1}^{k} \left( \frac{c_t}{N_{\text{row}} \times N_{\text{col}}} \right) \log\left( \frac{c_t}{N_{\text{row}} \times N_{\text{col}}} \right).
\end{equation}
The global-level information content of the parameters is calculated based on the difference in information entropy.
\begin{equation}
    V_G^{(A)} = f_{G}(|E(W_A) - E(W_B)|).
\label{eq:global_quality}
\end{equation}
$V_G^{(A)}$ represents the global-level model information content of the model $M_A$. 

\noindent \textit{Dual-Perspective \moduleAfull{}.}
The method of fusing global-level information content and local-level parameter uncertainty is as follows. 

\begin{equation}
    V^{(A)} = V_G^{(A)}*(1-\exp{(-V_G^{(A)}\times V_L^{(A)})})
\end{equation}
\begin{equation}
    V^{(A)} := \frac{\exp({V^{(A)}})}{\exp({V^{(A)}})+\exp({V^{(B)}})}
\end{equation}
The purpose of the above equation is to ensure that \( V^{(A)} + V^{(B)} = 1 \), in order to avoid the parameters of the integrated model being too large or too small.

\subsubsection{\moduleB{}}
After the assessment of the parameters' uncertainty, it is necessary to proceed with the splicing of the parameters.

\noindent \textit{Hard splicing.}
Hard splicing of parameters involves directly replacing parameters with lower compatibility with those of higher compatibility. $W$ represents the parameters of the spliced model.
\begin{equation}
\begin{aligned}
& W = W_A \odot V^{(A)} + W_B \odot V^{(B)}, \\
& \forall v^{(A)} \in V^{(A)},  v^{(A)} \in \{0, 1\}, \quad \\
& \forall v^{(B)} \in V^{(B)},  v^{(B)} \in \{0, 1\}.
\label{eq:hard_fusion}
\end{aligned}
\end{equation}

\noindent \textit{Soft splicing.}
While hard splicing, which selects low-compatibility parameters and discards high-compatibility ones, seems reasonable, it is important to note that low-compatibility parameters may not necessarily be suitable for the final fused model. Therefore, the direct transplantation of parameters could potentially lead to a decrease rather than an increase in the performance of the fused model. Therefore, a softer approach to splicing is worth exploring, namely soft splicing. $W$ represents the parameters of the spliced model.
\begin{equation}
\begin{aligned}
& W = W_A \odot V^{(A)} + W_B \odot V^{(B)}, \\
& \forall v^{(A)} \in V^{(A)}, 0 \leq v^{(A)} \leq 1, \quad \\
& \forall v^{(B)} \in V^{(B)}, 0 \leq v^{(B)} \leq 1.
\end{aligned}
\label{eq:soft_fusion}
\end{equation}

\noindent \textit{Generalization for \methodbrief{}.}
In order to better understand the generalization of \methodbrief{}, we take hard splicing as an example and use classical generalization theory to analyze the generalization boundary of the spliced model. We show that under some assumptions, the generalization boundary of spliced model $W$ can be bounded, see Appendix for more theoretical information.

\subsection{\methodbrief{} for Multiple Models}
\subsubsection{\moduleAfull{}} Here we introduce how to do parameter compatibility assessment based on multiple models.

\noindent \textit{Local-level \moduleA{}.} When the trained model is no longer just the two models, but rather a collection of $ n $ models, the set of the models parameter is $ \mathcal{M} = \{M_1, M_2, ..., M_n\} $, the set of the models parameter is $ \mathcal{W} = \{W_1, W_2, ..., W_n\} $,
the local-level uncertainty of all pairs can be regarded as a $n\times n$ matrix.
\begin{equation}
\begin{bmatrix}
V_L^{(1,1)} & V_L^{(1,2)} & \cdots & V_L^{(1,n)} \\
V_L^{(2,1)} & V_L^{(2,2)} & \cdots & V_L^{(2,n)} \\
\vdots & \vdots & \ddots & \vdots \\
V_L^{(n,1)} & V_L^{(n,2)} & \cdots & V_L^{(n,n)}
\end{bmatrix}
\label{eq:local_quality_matrix_multi}
\end{equation}
Then, following the formula above, we sum up the uncertainty of each row, and then we can obtain an uncertainty set $ \{V_L\} $ of length $ n $. $V_L^{(k)}$ represents the local-level uncertainty of the $k$-th model, \(i\) and \(j\) represent the row index and column index of the parameter matrix, respectively.
\begin{equation}
    \{ V_L^{(k)}[i,j] = \sum_{\substack{W_l \in \mathcal{W} \\ W_l \neq W_k}} f_L(|W_k[i,j] - W_l[i,j]|) \mid W_k \in \mathcal{W} \}
\label{eq:local_quality_final_multi}    
\end{equation}

\begin{table*}[!ht]
\renewcommand{\arraystretch}{1.035}
\centering
\resizebox{\textwidth}{!}{
\begin{tabular}{c|c|c|c|c|c|c|c|c|c|c|c}
\toprule[2pt]
 &  & \multicolumn{4}{c|}{\textbf{Metrics}} &  &  & \multicolumn{4}{c}{\textbf{Metrics}} \\ \cline{3-6} \cline{9-12}
\multirow{-2}{*}{\textbf{Datasets}} & \multirow{-2}{*}{\textbf{Methods}} & NDCG@5 & {HR@5} & NDCG@10 & HR@10 & \multirow{-2}{*}{\textbf{Datasets}} & \multirow{-2}{*}{\textbf{Methods}} & NDCG@5 & HR@5 & NDCG@10 & HR@10 \\
\midrule \midrule
 & Model A & {0.1137} & 0.1878 & {0.1561} & {0.3203} &  & {Model A} & \underline{0.3779} & \underline{0.5036} & \underline{0.4179} & 0.6273 \\
 & Model B & 0.1136 & 0.1888 & 0.1538 & 0.3143 &  & Model B & 0.3763 & 0.5034 & 0.4173 & \underline{0.6301} \\
 & Pruning & \underline{0.1250} & \underline{0.2129} & \underline{0.1833} & \underline{0.3936}  &  & Pruning & 0.3480 & 0.4727 & 0.3890 & 0.5998  \\
 & Ensemble & 0.1193 & 0.2038 & 0.1679 & 0.3534  &  & Ensemble & 0.3658 & 0.4882 & 0.4078 & 0.6181  \\
 & Averaging & 0.1081 & 0.1958 & 0.1682 & 0.3815 &  & Averaging & 0.3748 & 0.5030 & 0.4149 & 0.6269 \\
\multirow{-6}{*}{\makecell{\texttt{Beauty}}} & \cellcolor[HTML]{F2F2F2}\textbf{Ours} & \cellcolor[HTML]{F2F2F2}\textbf{0.1409} & \cellcolor[HTML]{F2F2F2}\textbf{0.2420} & \cellcolor[HTML]{F2F2F2}\textbf{0.1905} & \cellcolor[HTML]{F2F2F2}\textbf{0.3956}& \multirow{-6}{*}{\makecell{\texttt{Music}}} & \cellcolor[HTML]{F2F2F2}\textbf{Ours} & \cellcolor[HTML]{F2F2F2}\textbf{0.3805} & \cellcolor[HTML]{F2F2F2}\textbf{0.5064} & \cellcolor[HTML]{F2F2F2}\textbf{0.4212} & \cellcolor[HTML]{F2F2F2}\textbf{0.6326}\\
\midrule 
 & Model A & 0.4027 & 0.5260 & 0.4424 & 0.6490 &  & Model A & 0.4384 & 0.5897 & 0.4818 & 0.7208 \\
 & Model B & 0.4011 & 0.5255 & 0.4428 & 0.6544 &  & Model B & 0.4365 & 0.5901 & 0.4771 & 0.7158 \\
 & Pruning & 0.3756 & 0.4990 & 0.4190 & 0.6337  &  & Pruning & 0.3484 & 0.5014 & 0.3975 & 0.6534  \\
 & Ensemble & \underline{0.4089} & \underline{0.5322} & \underline{0.4503} & \underline{0.6600} &  & Ensemble & \underline{0.4464} & \underline{0.5982} & \textbf{0.4877} & \textbf{0.7257} \\
 & Averaging & 0.4015 & 0.5264 & 0.4428 & 0.6538 &  & Averaging & 0.4069 & 0.5674 & 0.4497 & 0.6994 \\
\multirow{-6}{*}{\makecell{\texttt{Book}}} & \cellcolor[HTML]{F2F2F2}\textbf{Ours} & \cellcolor[HTML]{F2F2F2}\textbf{0.4095} & \cellcolor[HTML]{F2F2F2}\textbf{0.5344} & \cellcolor[HTML]{F2F2F2}\textbf{0.4505} & \cellcolor[HTML]{F2F2F2}\textbf{0.6612}& \multirow{-6}{*}{\makecell{\texttt{Movielens}}} & \cellcolor[HTML]{F2F2F2}\textbf{Ours} & \cellcolor[HTML]{F2F2F2}\textbf{0.4480} & \cellcolor[HTML]{F2F2F2}\textbf{0.6001} & \cellcolor[HTML]{F2F2F2}\underline{0.4875} & \cellcolor[HTML]{F2F2F2}\underline{0.7219}\\
\bottomrule[2pt]
\end{tabular}
}
\caption{
Performance comparison on recommendation tasks when the pre-trained model is a static models. 
}
\label{tab:static}
\end{table*}
\begin{table}[!ht]
\centering
\renewcommand{\arraystretch}{1.035}
\resizebox{0.42\textwidth}{!}{
\begin{tabular}{c|c|c|c|c|c}
\toprule[2pt]
\multirow{2}{*}{\textbf{Datasets}} & \multirow{2}{*}{\textbf{Methods}} & \multicolumn{4}{c}{\textbf{Metric}} \\ \cline{3-6}
 &  & Acc & AUC & Recall & F1 \\
 \toprule \toprule
\multirow{6}{*}{\texttt{SST-2}} & Model A & 0.8429 & 0.8432 & 0.8429 & 0.8429 \\
     & Model B & 0.8440 & 0.8444 & 0.8440 & 0.8440 \\
 & Pruning & 0.8177 & 0.8111 & 0.8131 & 0.8106 \\
 & Ensemble & \underline{0.8532} & \underline{0.8530} & \underline{0.8532} & \underline{0.8532} \\
 & Averaging & 0.8406 & 0.8406 & 0.8406 & 0.8406 \\
 & \cellcolor[HTML]{F2F2F2}Ours & \cellcolor[HTML]{F2F2F2}\textbf{0.8544} & \cellcolor[HTML]{F2F2F2}\textbf{0.9161} & \cellcolor[HTML]{F2F2F2}\textbf{0.8853} & \cellcolor[HTML]{F2F2F2}\textbf{0.8853} \\
 \toprule 
\multirow{6}{*}{\texttt{RTE}} & Model A & 0.5884 & \underline{0.5959} & 0.5884 & 0.5815 \\
 & Model B & 0.5848 & 0.5854 & 0.5848 & 0.5851 \\
 & Pruning & 0.5271 & 0.5495 & 0.5523 & 0.5513 \\
 & Ensemble & \underline{0.6065} & \textbf{0.6091} & \underline{0.6065} & \underline{0.6061} \\
 & Averaging & 0.5740 & 0.5837 & 0.5740 & 0.5610 \\
 & \cellcolor[HTML]{F2F2F2}Ours & \cellcolor[HTML]{F2F2F2}\textbf{0.6245} & \cellcolor[HTML]{F2F2F2}0.5628 & \cellcolor[HTML]{F2F2F2}\textbf{0.6679} & \cellcolor[HTML]{F2F2F2}\textbf{0.6665} \\
\toprule[2pt]
\end{tabular}
}
\caption{
Performance comparison on language tasks. 
}
\label{tab:main_nlp}
\end{table}

\noindent \textit{Global-level \moduleAAA{}.}
Similarly to Equation~\ref{eq:local_quality_matrix_multi}, we  obtain the global-level information content matrix. Then, like Equation~\ref{eq:local_quality_final_multi}, we get a global-level information content set $ \{V_G\} $ of length $ n $. $V_G^{(k)}$ represents the global-level information content of the $k$-th model. Specifically, it can be formulated as follow,
\begin{equation}
    \{ V_G^{(k)} = \sum_{\substack{W_l \in \mathcal{W} \\ W_l \neq W_k}} f_G(|E(W_k) - E(W_l)|) \mid W_k \in \mathcal{W} \}
    \label{eq:global_quality_final_multi}
\end{equation}
\noindent \textit{Dual-Perspective \moduleAfull{}.}
After obtaining $ V_L $ and $ V_G $, we fuse the local-level uncertainty $ V_L $ and global-level information content $ V_G $ for each model $ M_k $ according to the following formula to get the model information content matrix $ V^{(k)} $.
\begin{equation}
    \{ V^{(k)} = V_G^{(k)}*(1-e^{-V_G^{(k)}\times V_L^{(k)}}) \mid M_k \in \mathcal{M} \}
\end{equation}
Then we proceed by normalizing the model information content.
\begin{equation}
    \{ V^{(k)} := \frac{\exp(V^{(k)})}{\sum_{\substack{M_l \in \mathcal{M}}} \exp(V^{(l)})} \mid M_k \in \mathcal{M} \}
\end{equation}
The purpose of the above equation is to ensure that \( V^{(A)} + V^{(B)} = 1 \), in order to avoid the parameters of the integrated model being too large or too small.

\subsubsection{\moduleB{}}
The \moduleB{} methods also include two types: hard splicing and soft splicing. In terms of \moduleB{}, we extend the formulas based on Equations~\ref{eq:hard_fusion} and \ref{eq:soft_fusion} to a broader range of applications.
\begin{equation}
\begin{aligned}
 W = \sum_{W_k \in \mathcal{W}} W_k \odot V^{(k)}, \forall v^{(k)} \in  V^{(k)}, 0 \leq  V^{(k)} \leq 1.
\end{aligned}
\end{equation}
\begin{table*}[!ht]
\renewcommand{\arraystretch}{1.035}
\centering
\resizebox{\textwidth}{!}{
\begin{tabular}{c|c|c|c|c|c|c|c|c|c|c|c}
\toprule[2pt]
 &  & \multicolumn{4}{c|}{\textbf{Metrics}} &  &  & \multicolumn{4}{c}{\textbf{Metrics}} \\ \cline{3-6} \cline{9-12}
\multirow{-2}{*}{\textbf{Datasets}} & \multirow{-2}{*}{\textbf{Methods}} & \multicolumn{1}{c|}{NDCG@5} & \multicolumn{1}{c|}{HR@5} & \multicolumn{1}{c|}{NDCG@10} & \multicolumn{1}{c|}{HR@10} & \multirow{-2}{*}{\textbf{Datasets}} & \multirow{-2}{*}{\textbf{Methods}} & \multicolumn{1}{c|}{NDCG@5} & \multicolumn{1}{c|}{HR@5} & \multicolumn{1}{c|}{NDCG@10} & \multicolumn{1}{c}{HR@10} \\
\midrule \midrule
 & Model A & 0.1137 & 0.1878 & 0.1561 & 0.3203  &  & Model A & \underline{0.3779} & 0.5036 & \underline{0.4179} & 0.6273  \\
 & Model B (G) & \underline{0.2046} & \underline{0.2982} & \underline{0.2354} & 0.3926 & & Model B (G) & 0.3760 & \underline{0.5082} & 0.4173 & \underline{0.6356}  \\
 & Pruning & 0.1370 & 0.2229 & 0.1862 & 0.3775  &  & Pruning & 0.3150 & 0.4464 & 0.3614 & 0.5901  \\
 & Ensemble & 0.1901 & 0.2751 & 0.2181 & 0.3604  &  & Ensemble & 0.3728 & 0.5029 & 0.4136 & 0.6292  \\
 & Averaging & 0.1613 & 0.2620 & 0.2064 & \underline{0.4016}  &  & Averaging & 0.3747 & 0.5008 & 0.4159 & 0.6278  \\
\multirow{-6}{*}{\texttt{Beauty}} & \cellcolor[HTML]{F2F2F2}Ours & \cellcolor[HTML]{F2F2F2}\textbf{0.2296} & \cellcolor[HTML]{F2F2F2}\textbf{0.3183} & \cellcolor[HTML]{F2F2F2}\textbf{0.2573} & \cellcolor[HTML]{F2F2F2}\textbf{0.4046}  & \multirow{-6}{*}{\texttt{Music}} & \cellcolor[HTML]{F2F2F2}Ours & \cellcolor[HTML]{F2F2F2}\textbf{0.3929} & \cellcolor[HTML]{F2F2F2}\textbf{0.5201} & \cellcolor[HTML]{F2F2F2}\textbf{0.4332} & \cellcolor[HTML]{F2F2F2}\textbf{0.6444}  \\
\midrule 
 & Model A & 0.4027 & 0.5260 & 0.4424 & 0.6490  &  & Model A & 0.4384 & 0.5897 & 0.4818 & 0.7228  \\
 & Model B (G) & 0.3959 & 0.5201 & 0.4374 & 0.6484  &  & Model B (G) & 0.4556 & 0.6205 & 0.4964 & 0.7463  \\
 & Pruning & 0.3463 & 0.4739 & 0.3904 & 0.6101  &  & Pruning & 0.3335 & 0.4963 & 0.3857 & 0.6580  \\
 & Ensemble & 0.4078 & \underline{0.5315} & 0.4472 & 0.6536  &  & Ensemble & 0.4665 & 0.6215 & 0.5038 & 0.7361  \\
 & Averaging & \underline{0.4080} & 0.5313 & \underline{0.4487} & \underline{0.6571}  &  & Averaging & \underline{0.4680} & \underline{0.6273} & \underline{0.5072} & \underline{0.7480}  \\
\multirow{-6}{*}{\texttt{Book}} & \cellcolor[HTML]{F2F2F2}Ours & \cellcolor[HTML]{F2F2F2}\textbf{0.4160} & \cellcolor[HTML]{F2F2F2}\textbf{0.5436} & \cellcolor[HTML]{F2F2F2}\textbf{0.4574} & \cellcolor[HTML]{F2F2F2}\textbf{0.6716}  & \multirow{-6}{*}{\texttt{Movielens}}   & \cellcolor[HTML]{F2F2F2}Ours & \cellcolor[HTML]{F2F2F2}\textbf{0.4811} & \cellcolor[HTML]{F2F2F2}\textbf{0.6315} & \cellcolor[HTML]{F2F2F2}\textbf{0.5211} & \cellcolor[HTML]{F2F2F2}\textbf{0.7535}  \\
\bottomrule[2pt]
\end{tabular}
}
\caption{
Performance comparison on recommendation task when pre-trained models include both static and dynamic models. 
}
\label{tab:dynamic}
\end{table*}
\section{Experiments}
\label{sec:experiments}

\subsection{Experimental Setup}
\subsubsection{Datasets}
\begin{sloppypar}
In recommendation tasks, we evaluated our method on 4 widely used datasets \texttt{Amazon-Beauty (Beauty)}, \texttt{Douban-Book (Book)}, \texttt{Douban-Music (Music)}, \texttt{Movielens-1M (Movielens)}. 
In language tasks, we evaluated our method on 2 widely used datasets \texttt{RTE}, and \texttt{SST-2}. Since the usage of the aforementioned language datasets is relatively conventional and straightforward, we will not provide an extensive introduction to the preprocessing of these datasets here.
Due to the more complex data distribution in recommendation tasks, the distribution shift of the data is more pronounced and rapid. Therefore, we prefer to conduct experiments on recommendation tasks.
\end{sloppypar}

\subsubsection{Baselines}
\label{subsec:experiment_baseline}
To evaluate the effectiveness of our method, we selected baselines from the following categories: \textbf{{Static Recommendation Models.}}
    \textit{DIN}~\cite{ref:din}, \textit{GRU4Rec}~\cite{ref:gru4rec}, and \textit{SASRec}~\cite{ref:sasrec} are three widely used  sequential recommendation methods. \textbf{{Dynamic Recommendation Framework.}}
    \textit{DUET}~\cite{ref:duet} is a framework that can generate  model parameters for the static model during inference based on the sample. \textbf{{Language Models.}}
    \textit{Roberta}~\cite{ref:liu2019roberta} is a widely used language method, which is improved based on \textit{Bert}~\cite{ref:devlin2018bert}. \textbf{{Methods to address incompatible parameter issue.}}
    \textit{Model Pruning}~\cite{ref:pruning_guo2020multi,ref:pruning_han2015learning} and \textit{Output Ensemble}~\cite{ref:ensemble_zhou2021domain_adaptation} address the incompatible parameter issue to some extent by cutting off unimportant connections in the neural network and integrating inference results, respectively. \textbf{{Other model fusion methods.}}
    \textit{Parameter Average} averages parameters of multiple pretrained models.

\subsubsection{Evaluation Metrics} 
In the experiments, we use the widely adopted \textit{Acc} (Accuracy), \textit{AUC}, \textit{Recall}, \textit{F1} for language tasks, and use widely adopted \textit{AUC}, \textit{UAUC}, \textit{NDCG}, and \textit{HR} (HitRate) as the metrics to evaluate model performance. 

\subsection{Experimental Results}
To facilitate performance comparison, the \textbf{best} value is highlighted in bold, while the \underline{second best} value is underlined. Some tables only highlight the maximum value.

\subsubsection{\methodbrief{} for Global Incompatible Parameters}
\begin{sloppypar}
Table~\ref{tab:static} and \ref{tab:main_nlp} summarize the comparison of our method and other methods to address incompatible parameter issue based on static recommendation models. 
The $1$-st and $2$-nd rows, Model $M_A$ and Model $M_B$, are pre-trained models with different initial conditions. The $3$-rd row, Pruning, is a model pruning based on Model $M_A$. The $4-$th row is to ensemble the outputs by the Model $M_A$ and Model $M_B$. The $5-$th row is to average the parameters of the Model $M_A$ and Model $M_B$. The $6-$th row is our method, which can do knowledge integration based on the compatibility. To further validate the effectiveness of our method, we compared it with other model fusion methods and output fusion methods. 
From this table, we have the following conclusions: (1) Pruning often performs worse than the original model because the purpose of pruning is not to enhance the capability of incompatible parameters but to lighten the model. In the process of removing incompatible parameters, it sometimes also cuts well-trained parameters. (2) In some cases, our method achieves significant improvements compared to all these baselines. It is worth noting that the inference cost of the output ensemble is $n$ times that of our method. The reason why result ensemble has an Inference cost of $n\times$ is because it requires multiple models (in this case, $\#\text{models}=2$) to perform inference separately and then integrate the results. Nevertheless, our method achieves better performance than all baselines in almost all cases.
Experimental results show that our method significantly improves model performance with the same resource consumption. This indicates that our approach is effective in optimizing parameters that are incompatible with the global data distribution.
\end{sloppypar}

\subsubsection{\methodbrief{} for Individual Incompatible Parameters}

Table~\ref{tab:dynamic} compares the performance of parameter fusion based on dynamic recommendation models (model $M_B$ (G)). Unlike the parameter fusion methods for static models, the parameters of $M_B$ (G) are dynamically generated by a hypernetwork based on the input samples. Both dynamic and static models use the same base model for parameter fusion. Across all datasets and metrics, parameter fusion significantly outperforms other baselines. Additionally, Comparing Table~\ref{tab:static} and Table~\ref{tab:dynamic}, we find that parameter fusion between dynamic and static models achieves a more significant performance improvement over baseline models than parameter fusion between static models. This is because the parameters of dynamic models are adaptively generated in an unsupervised manner based on input samples. Although the performance of dynamic models significantly surpasses that of static models, they also exhibit greater instability. This instability can be compensated by incorporating parameters from static models. Experimental results demonstrate that this method is highly effective in optimizing parameters that are incompatible with individual data distributions.

\subsubsection{\methodbrief{} for Initialization and Re-training}
\begin{table}[!ht]
\renewcommand{\arraystretch}{1.035}
\centering
\resizebox{0.85\linewidth}{!}{
\begin{tabular}{c|c|c|c|c}
\toprule[2pt]
\multirow{2}{*}{\textbf{Models}} & \multicolumn{4}{c}{\textbf{Metrics}} \\ \cline{2-5}
 & NDCG@5 & HR@5 & NDCG@10 & HR@10 \\
\midrule\midrule
Base & 0.3779 & 0.5036 & 0.4179 & 0.6273 \\
Ours & 0.3823 & 0.5071 & 0.4221 & 0.6302 \\
Ours (F) & \textbf{0.3831} & \textbf{0.5091} & \textbf{0.4223} & \textbf{0.6304} \\
\bottomrule[2pt]
\end{tabular}
}
\caption{\method{} for model initialization and re-training. 
}
\label{tab:analysis_finetune}
\end{table}
After the model parameters are integrated, the resulting model can be directly used for inference, or it can serve as a better initial parameter set, allowing the model to be further trained on this basis for improved results. To verify the initial performance of the model after \methodbrief{}, we trained the model again post-integrating. As shown in Table~\ref{tab:analysis_finetune}, We found that the integrated model only needs one epoch to converge and achieve better results than the parameter integrated model and other baselines. This also demonstrates the benefits of \methodbrief{} for model initialization.
\section{Conclusion}
\label{sec:conclusion}

We introduced \methodbrief{}, a novel method to optimize incompatible parameters in deep neural networks by leveraging the complementary strengths of different pretrained models, without adding extra parameters. Our experiments on recommendation and language tasks demonstrate that \methodbrief{} effectively enhances model performance without increasing inference costs, offering a promising approach for deploying more robust and efficient models.
\section*{Acknowledgments}
This work was supported by 2030 National Science and Technology Major Project (2022ZD0119100), Scientific Research Fund of Zhejiang Provincial Education Department (Y202353679), National Natural Science Foundation of China (No. 62402429, 62376243, 62441605, 62037001), the Key Research and Development Program of Zhejiang Province (No. 2024C03270), ZJU Kunpeng$\&$Ascend Center of Excellence, the Key Research and Development Projects in Zhejiang Province (No.2024C01106), Zhejiang University Education Foundation Qizhen Scholar Foundation, the Starry Night Science Fund at Shanghai Institute for Advanced Study (Zhejiang University). This work was also supported by Ant group.
\bibliography{aaai25}
\appendix

\section{Appendix}
\label{sec:appendix}

\subsection{Supplementary Methodology}
\subsubsection{Pseudo Code}
Algorithm~\ref{alg:pseudo_code} shows the pseudo code of Compatibility-aware Knowledge Integration.
\renewcommand{\algorithmicrequire}{\textbf{Input:}}  
\renewcommand{\algorithmicrequire}{\textbf{Initialization:}}
\begin{algorithm}[!h]
\begin{flushleft}
  \caption{Optimize Incompatible Parameters through Compatibility-aware Knowledge Integration}

\textbf{Variable}: Data $\mathcal{X}=\{x\}$, Local-level Parameter Uncertainty Set $\mathcal{V_L} = \{V_L \mid M_k \in \mathcal{M}\}$, Global-level Model Information Content Set $\mathcal{V_G} = \{V_G \mid M_k \in \mathcal{M}\}$, Parameter Compatibility Set $\mathcal{V} = \{V \mid M_k \in \mathcal{M}\}$, Pretrained Model Set $\mathcal{M}$, Integrated Model $M$, Local-level Parameter Uncertainty Network $f_L$, Global-level Model Information Content Network $f_G$.

\textbf{Module 1:}~\colorbox{gray!30}{$\rhd$~\emph{Parameter Compatibility Assessment}}
    \resizebox{0.46\textwidth}{!}{
    \begin{tcolorbox}[sharp corners, colframe=gray!80!white, colback=white, boxrule=0.5mm, left=0pt, right=0pt, top=0pt, bottom=0pt, boxsep=5pt]
    \begin{algorithmic}
    \State \textbf{Target}: $\mathcal{X}, \mathcal{M} \mapsto \mathcal{V}$
    \State \textbf{Input}: $\mathcal{X}, \mathcal{M}$
    \State \textbf{Intermediate}: $\mathcal{X}, \mathcal{M} \mapsto \mathcal{V_L}, \mathcal{V_G} \mapsto \mathcal{V}$
    \State \textbf{Output}: $\mathcal{V}$
    \end{algorithmic}
    \end{tcolorbox}
    }

\textbf{Module 2:}~\colorbox{gray!30}{$\rhd$~\emph{Parameter Splicing}}
\resizebox{0.46\textwidth}{!}{
    \begin{tcolorbox}[sharp corners, colframe=gray!80!white, colback=white, boxrule=0.5mm, left=0pt, right=0pt, top=0pt, bottom=0pt, boxsep=5pt]
    \begin{algorithmic}
    \State \textbf{Target}: $\mathcal{V}, \mathcal{M} \mapsto M$
    \State \textbf{Input}: $\mathcal{V}, \mathcal{M}$
    \State \textbf{Output}: $M$
    \end{algorithmic}
    \end{tcolorbox}
    }

\textbf{Overview:}~\colorbox{gray!30}{$\rhd$~\emph{Training Procedure}}\\
    \noindent \textbf{Repeat}\\
    \!\!\quad \textbf{If} {$f_L$ and $f_G$ have not yet been well-trained} \\
     {  
        \begin{algorithmic}
        \State \!\!\!\!\quad I. Do forward propagation to get $\mathcal{V_L}$ as Eq.\ref{eq:local_quality_final_multi}, $\mathcal{V_G}$ as Eq.\ref{eq:global_quality_final_multi}.
        \State \!\!\!\!\quad II. Choose a splicing method from hard splicing~Eq.\ref{eq:hard_fusion} and soft splicing~Eq.\ref{eq:soft_fusion}.
        \State \!\!\!\!\quad III. Train as $\mathop{\rm min}_{f_L, f_G}\mathcal{L} = \sum\nolimits_{x\in\mathcal{X},y\in\mathcal{Y}}l_{\text{}}(y, \hat{y}=M(x))$ 
        \end{algorithmic}
           }
    \noindent \textbf{Until} {\textit{Convergence}}
\label{alg:pseudo_code}
\end{flushleft}
\end{algorithm}
\subsubsection{Training and Inference Procedure}
\noindent \textit{Training Procedure.}
During training, in the dual-model \methodbrief{}, what we need to do is obtain the optimal $ V=\{V^{(A)}, V^{(B)}\} $ to minimize the loss $ \mathcal{L} $. When extending to multi-model \methodbrief{}, our goal is similar. Assuming the number of models is $ N $, the optimization objective is to find the optimal compatibility set $ V = \{V^{(k)}\}_{k=1}^{n} $ to minimize $ \mathcal{L} $, that is, $\argminA_{V = \{V^{(k)}\}_{k=1}^{n}} \mathcal{L}$.

\noindent \textit{Inference Procedure.}
After the splicing of model parameters, the parameters $ W = \{W_k\}_{k=1}^{n} $ of $ n $ models are combined into the parameters of one model $ W $, and then this model can perform inference just like a regular recommendation model, that is, $\hat{y} = Wx$.

\begin{table}[!ht]
\centering
\renewcommand{\arraystretch}{1.1}
\resizebox{1\linewidth}{!}{
\begin{tabular}{c|c|c|c|c|c}
\toprule[2pt]
\multicolumn{2}{c|}{\makecell{\textbf{Assessment Methods}}} & \multicolumn{4}{c}{\textbf{Metrics}} \\ 
\hline
Local Uncertainty & Global Information Content & AUC & UAUC & NDCG@5 & HR@5  \\ \midrule \midrule
\XSolidBrush & \XSolidBrush & 0.8557 & 0.8534 & 0.3779 & 0.5036  \\
\Checkmark & \XSolidBrush & 0.8554 & 0.8534 & 0.3797 & 0.5057  \\
\XSolidBrush & \Checkmark &  \underline{0.8582} & \underline{0.8558} & \underline{0.3803} & \underline{0.5064}  \\
\Checkmark & \Checkmark & \textbf{0.8585} & \textbf{0.8560} & \textbf{0.3823} & \textbf{0.5071}  \\
\bottomrule[2pt]
\end{tabular}
}
\caption{Results of the ablation studies \textit{w.r.t.} the assessment methods. The best performance is highlighted in bold while the second best performance is underlined.}
\label{tab:ablation_assessment}
\end{table}

\subsection{Supplementary Experiments}
\subsubsection{Evaluation Metrics}
We use \textit{Cost} to measure the model's consumption of resources. The calculation of \textit{Cost} is the inference time of the method divided by the inference time of the base model, rounded to one decimal place.

\subsubsection{The Effect of \moduleAfull{} Methods}
\begin{sloppypar}
To evaluate the impact of local uncertainty and global information content, we conducted ablation experiments on the \moduleAfull{} method. As shown in Table~\ref{tab:ablation_assessment}, there are 4 rows of experimental results. The results show: (1) The best performance is achieved when both local uncertainty and global information content are fused into a dual-perspective parameter compatibility. \textit{In summary, the dual-perspective \moduleAfull{} achieves the best $\text{NDCG@5}=0.3823$ and $\text{HR@5}=0.5071$, which is a significant improvement compared to the best one-perspective \moduleAfull{} $\text{NDCG@5}=0.3803$ and $\text{HR@5}=0.5064$.} (2) When only one type of \moduleAfull{} method is used, the global information content shows better performance than the local uncertainty, indicating that global information content is more important. (3) Both dual perspective or one perspective show better performance than no \moduleAfull{}. 
\end{sloppypar}

\subsubsection{The Effect of \moduleB{} Methods}
\moduleB{} is including hard splicing and soft splicing. Table~\ref{tab:ablation_fusion}, shows the comparison of hard splicing and soft splicing.
The experimental results indicate that soft splicing is a superior method compared to hard splicing.

\begin{table}[!h]
\centering
\renewcommand{\arraystretch}{1.05}
\resizebox{0.36\textwidth}{!}{
\begin{tabular}{c|c|c|c|c}
\toprule[2pt]
\multirow{2}{*}{\textbf{Fusion Methods}} & \multicolumn{4}{c}{\textbf{Metrics}} \\ \cline{2-5}
 & AUC & UAUC & NDCG@5 & HR@5 \\
\midrule \midrule
Hard splicing & 0.8561 & 0.8532 & \textbf{0.3806} & 0.5047 \\
Soft splicing & \textbf{0.8581} & \textbf{0.8558} & 0.3805 & \textbf{0.5064} \\
\bottomrule[2pt]
\end{tabular}
}
\caption{Results of the ablation studies \textit{w.r.t.} the splicing methods. The best performance is highlighted.}
\label{tab:ablation_fusion}
\end{table}

\begin{table*}[!ht]
\renewcommand{\arraystretch}{1.2}
\resizebox{\textwidth}{!}{
\begin{tabular}{c|c|c|c|c|c|c|c|c|c|c|c|c|c}
\toprule[2pt]
 &  &  & \multicolumn{4}{c|}{\textbf{Metrics}} &  &  &  & \multicolumn{4}{c}{\textbf{Metrics}} \\
\cline{4-7} \cline{11-14}
\multirow{-2}{*}{\textbf{Datasets}} & \multirow{-2}{*}{\textbf{\makecell{Base\\Models}}} & \multirow{-2}{*}{\textbf{Methods}} & NDCG@5 & HR@5 & NDCG@10 & HR@10 & \multirow{-2}{*}{\textbf{Datasets}} & \multirow{-2}{*}{\textbf{\makecell{Base\\Models}}} & \multirow{-2}{*}{\textbf{Methods}} & NDCG@5 & HR@5 & NDCG@10 & HR@10 \\ \midrule \midrule
 &  & Model A & \underline{0.4027} & \underline{0.5260} & 0.4424 & 0.6490 &  &  & Model A & \underline{0.3779} & \underline{0.5036} & \underline{0.4179} & 0.6273 \\
 &  & Model B & 0.4011 & 0.5255 & \underline{0.4428} & \underline{0.6544} &  &  & Model B & 0.3763 & 0.5034 & 0.4173 & \underline{0.6301} \\
 & \multirow{-3}{*}{DIN} & \cellcolor[HTML]{F2F2F2}Ours & \cellcolor[HTML]{F2F2F2}\textbf{0.4095} & \cellcolor[HTML]{F2F2F2}\textbf{0.5344} & \cellcolor[HTML]{F2F2F2}\textbf{0.4505} & \cellcolor[HTML]{F2F2F2}\textbf{0.6612} &  & \multirow{-3}{*}{DIN} & \cellcolor[HTML]{F2F2F2}Ours & \cellcolor[HTML]{F2F2F2}\textbf{0.3805} & \cellcolor[HTML]{F2F2F2}\textbf{0.5064} & \cellcolor[HTML]{F2F2F2}\textbf{0.4212} & \cellcolor[HTML]{F2F2F2}\textbf{0.6326} \\ 
 \cline{2-7} \cline{9-14}
 &  & Model A & \underline{0.4094} & \underline{0.5329} & \underline{0.4525} & \underline{0.6658} &  &  & Model A & 0.3677 & 0.4988 & 0.4109 & 0.6321 \\
 &  & Model B & 0.4059 & 0.5303 & 0.4481 & 0.6602 &  &  & Model B & \underline{0.3718} & \underline{0.5024} & \underline{0.4150} & \textbf{0.6360} \\
 & \multirow{-3}{*}{GRU4Rec} & \cellcolor[HTML]{F2F2F2}Ours & \cellcolor[HTML]{F2F2F2}\textbf{0.4100} & \cellcolor[HTML]{F2F2F2}\textbf{0.5350} & \cellcolor[HTML]{F2F2F2}\textbf{0.4527} & \cellcolor[HTML]{F2F2F2}\textbf{0.6669} &  & \multirow{-3}{*}{GRU4Rec} & \cellcolor[HTML]{F2F2F2}Ours & \cellcolor[HTML]{F2F2F2}\textbf{0.3756} & \cellcolor[HTML]{F2F2F2}\textbf{0.5091} & \cellcolor[HTML]{F2F2F2}\textbf{0.4162} & \cellcolor[HTML]{F2F2F2}\underline{0.6347} \\ 
 \cline{2-7} \cline{9-14}
 &  & Model A & \underline{0.4033} & {0.5262} & 0.4445 & 0.6548 &  &  & Model A & 0.3734 & 0.5038 & 0.4160 & \underline{0.6358} \\ 
 &  & Model B & 0.4031 & \underline{0.5266} & \underline{0.4451} & \underline{0.6565} &  &  & Model B & \underline{0.3772} & \underline{0.5090} & \underline{0.4182} & 0.6357 \\
\multirow{-9}{*}{Books} & \multirow{-3}{*}{SASRec} & \cellcolor[HTML]{F2F2F2}Ours & \cellcolor[HTML]{F2F2F2}\textbf{0.4043} & \cellcolor[HTML]{F2F2F2}\textbf{0.5276} & \cellcolor[HTML]{F2F2F2}\textbf{0.4468} & \cellcolor[HTML]{F2F2F2}\textbf{0.6573} & \multirow{-9}{*}{Music} & \multirow{-3}{*}{SASRec} & \cellcolor[HTML]{F2F2F2}Ours & \cellcolor[HTML]{F2F2F2}\textbf{0.3799} & \cellcolor[HTML]{F2F2F2}\textbf{0.5111} & \cellcolor[HTML]{F2F2F2}\textbf{0.4221} & \cellcolor[HTML]{F2F2F2}\textbf{0.6416} \\
\bottomrule[2pt]
\end{tabular}
}
\caption{The impact of choosing different small models on performance. 
}
\label{tab:different_models}
\end{table*}
\begin{figure}[!ht]
    \centering
        \includegraphics[width=0.475\textwidth]{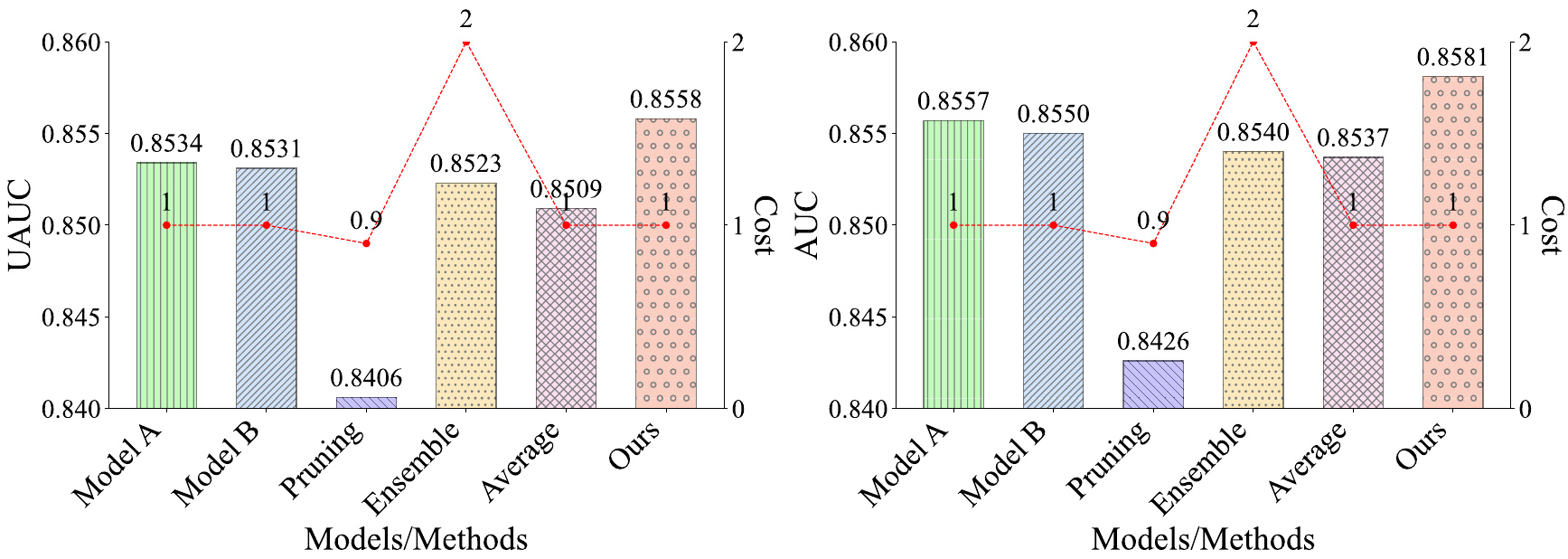}
        \caption{Performance comparison of the proposed method and baselines on recommendation task when the pre-trained model is a static models.}
    \label{fig:main_table}
\end{figure}
\begin{figure}[!ht]
\centering
        \includegraphics[width=0.475\textwidth]{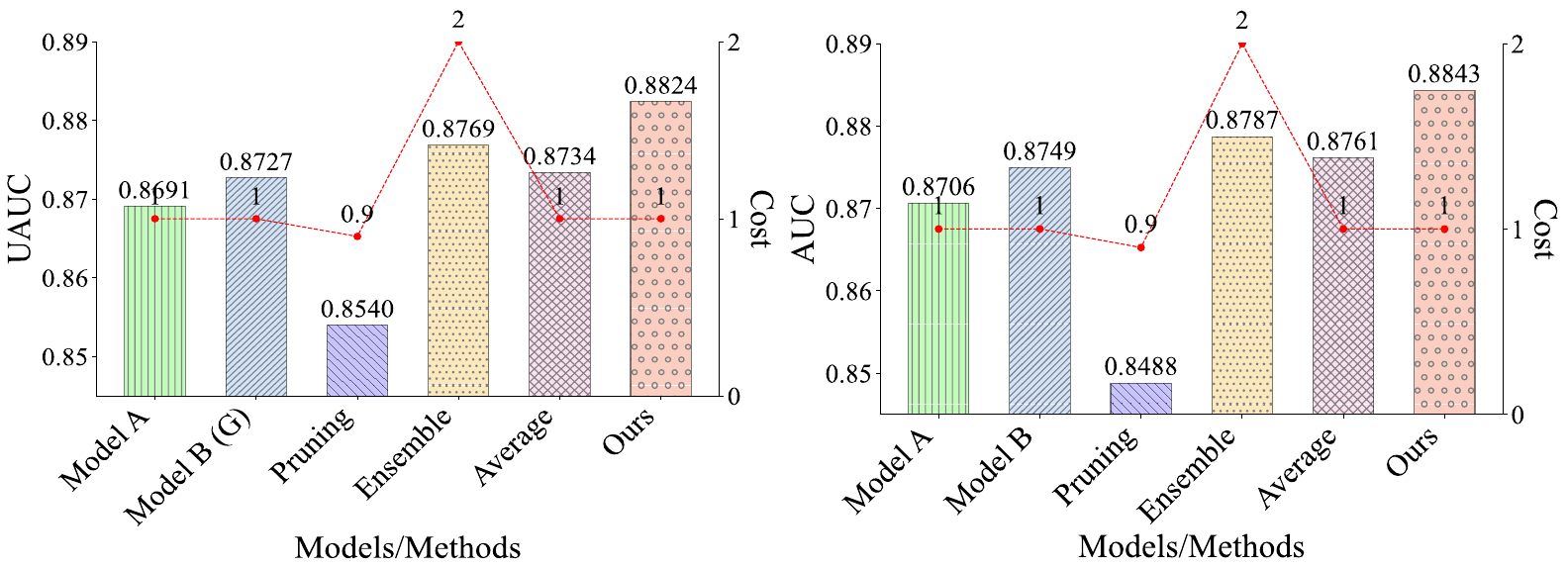}
        \caption{Performance comparison of the proposed method and baselines on recommendation task when pre-trained models include both static and dynamic models.}
    \label{fig:main_table_dynamic}
\end{figure}
Figure~\ref{fig:main_table} and \ref{fig:main_table_dynamic} shows the comparison of our method with baselines on  UAUC, AUC, and Cost. Each bar in the figure represents UAUC or AUC, while the line represents Cost.
All these tables and figures demonstrate that our method improved performance without increasing the cost.


\subsubsection{The Impact of the Different Base Models}
Table~\ref{tab:different_models} summarizes the performance comparison of our method, \methodbrief{}, with pre-trained models and Pruning on the Douban Book and Douban Music datasets when different small models are used. 
The results in the table show that our method is not limited by the choice of recommendation model, and \methodbrief{} demonstrates significant performance improvements across all recommendation models. For instance, when the GRU4Rec recommendation model is used, on the Douban Music dataset, it improves from the best pre-trained model $\text{NDCG@5}=0.3718$ to $\text{NDCG@5}=0.3756$, and $\text{HR@5}=0.5024$ to $\text{HR@5}=0.5091$.

\begin{figure}[!h]
    \centering
    \includegraphics[width=0.43\textwidth]{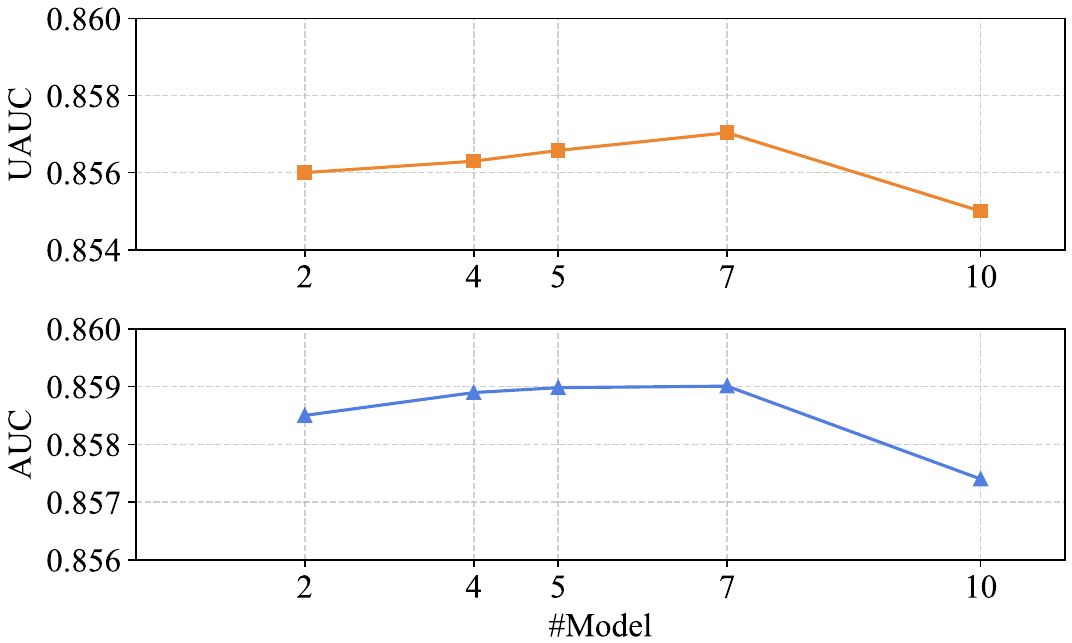}
    \caption{The impact of the number of integrated models.}
    \label{fig:analysis_number}
\end{figure}

\subsubsection{The Impact of the Number of Integrated Models}
As shown in Figure~\ref{fig:analysis_number}, 
we expanded model integrating from dual models to multiple models. The experimental results show: the best performance is achieved when the number of models is 7.
When the number of models is less than 7, the performance of the integrated model increases with the increase in the number of pretrained models. This is because, within a certain range, integrating of multiple model can more effectively leverage the knowledge learned from other models, optimizing the redundant parameters of the current model. However, an increase in the number of pretrained models leads to an increase in the size of the integrated network, and the available data volume can no longer support the training of a larger integrated network. Therefore, when the number of models exceeds 7, the performance of the integrated model rapidly declines with the increase in the number of pretrained models.

\subsubsection{Hyperparameters and Training Schedules}
\label{sec:appendix_implementation_detail}
We summarize the hyperparameters and training schedules of the datasets used in the experiments in Table~\ref{tab:hyperparameters_and_training_schedule}.
\begin{table}[!ht]
\centering
\resizebox{0.43\textwidth}{!}{
    \begin{tabular}{c|c|c}
    \toprule[2pt]
    \textbf{Dataset} & \textbf{Parameters} & \textbf{Setting} \\ 
    \midrule
    \midrule
    \multirow{5}{*}{\makecell[c]{SST-2\\RTE}} & GPU & Tesla A100 \\ \cline{2-3}
    \multirow{8}{*}{} & Optimizer & Adam\\ \cline{2-3}
    \multirow{8}{*}{} & \makecell[c]{Learning rate} & 0.001\\ \cline{2-3}
    \multirow{8}{*}{} & \makecell[c]{Batch size} & 32 \\ \cline{2-3}
    \multirow{8}{*}{} & \makecell[c]{the Dimension of embedding} & 1×514 \\ 
    \midrule
    \multirow{6}{*}{\makecell[c]{Amazon Beauty\\Douban Book\\Douban Music\\Movielens-1M}} & GPU & Tesla A100 \\ \cline{2-3}
    \multirow{8}{*}{} & Optimizer & Adam\\ \cline{2-3}
    \multirow{8}{*}{} & \makecell[c]{Learning rate} & 0.001\\ \cline{2-3}
    \multirow{8}{*}{} & \makecell[c]{Batch size} & 1024 \\ \cline{2-3}
    \multirow{8}{*}{} & \makecell[c]{Sequence length} & 10 \\ \cline{2-3}
    \multirow{8}{*}{} & \makecell[c]{the Dimension of embedding} & 1×32 \\ 
     \bottomrule[2pt]
    \end{tabular}
   }
    \caption{Hyperparameters of CKI.}
    \label{tab:hyperparameters_and_training_schedule}
\end{table}

\end{document}